\documentclass[conference]{IEEEtran}
\IEEEoverridecommandlockouts
\usepackage{cite}
\usepackage{amsmath,amssymb,amsfonts}
\usepackage{algorithmic}
\usepackage{graphicx}
\usepackage{textcomp}
\usepackage{xcolor}
\usepackage{booktabs,caption}
\usepackage{threeparttable}
\def\BibTeX{{\rm B\kern-.05em{\sc i\kern-.025em b}\kern-.08em
    T\kern-.1667em\lower.7ex\hbox{E}\kern-.125emX}}
\begin{document}

\title{AV-MaskEnhancer: Enhancing Video Representations through Audio-Visual Masked Autoencoder\\

\thanks{$^*$ These authors contributed equally to this work.}
}

\author{\IEEEauthorblockN{Xingjian Diao$^*$}
\IEEEauthorblockA{\textit{Department of Computer Science} \\
\textit{Dartmouth College}\\
Hanover, US \\
xingjian.diao.gr@dartmouth.edu}
\and
\IEEEauthorblockN{Ming Cheng$^*$}
\IEEEauthorblockA{\textit{Department of Computer Science} \\
\textit{Dartmouth College}\\
Hanover, US \\
ming.cheng.gr@dartmouth.edu}
\and
\IEEEauthorblockN{Shitong Cheng}
\IEEEauthorblockA{\textit{Department of Computer Science} \\
\textit{Dartmouth College}\\
Hanover, US \\
shitong.cheng.gr@dartmouth.edu}
}

\maketitle

\begin{abstract}
Learning high-quality video representation has shown significant applications in computer vision and remains challenging. Previous work based on mask autoencoders such as ImageMAE \cite{he2022masked} and VideoMAE \cite{tong2022videomae} has proven the effectiveness of learning representations in images and videos through reconstruction strategy in the visual modality. However, these models exhibit inherent limitations, particularly in scenarios where extracting features solely from the visual modality proves challenging, such as when dealing with low-resolution and blurry original videos. Based on this, we propose AV-MaskEnhancer for learning high-quality video representation by combining visual and audio information. Our approach addresses the challenge by demonstrating the complementary nature of audio and video features in cross-modality content. Moreover, our result of the video classification task on the UCF101 \cite{soomro2012ucf101} dataset {outperforms the existing work and reaches the state-of-the-art}, with a {top-1 accuracy} of 98.8\% and a {top-5 accuracy} of 99.9\%.
\end{abstract}
\newcommand\blfootnote[1]{%
  \begingroup
  \renewcommand\thefootnote{}\footnote{#1}%
  \addtocounter{footnote}{-1}%
  \endgroup
}

\blfootnote{2023 IEEE 35th International Conference on Tools with Artificial Intelligence (ICTAI)}

\section{Introduction}
\label{introduction}
Transformer\cite{vaswani2017attention} and Vision Transformer (ViT) \cite{dosovitskiy2020image} have brought significant improvements in latent representation in natural language processing and computer vision, respectively. In computer vision, multiple works \cite{he2022masked, tong2022videomae, gong2022contrastive} have proven the effectiveness of ViT as the encoder to extract deep features from large-scale image/video datasets. Unlike complicated and inefficient CNN-based structures \cite{xie2020self, hu2020dasgil, kolesnikov2020big}, ViT splits the original image into multiple patches, embeds these patches linearly, and utilizes an attention mechanism to capture long-range relationships and dependencies between them, allowing the model to focus on the important features and speed up the computation. 
Based on this, lots of recent studies use ViT as the feature encoder.
Recently, the ViT-based single-modal Masked Autoencoders (MAE) \cite{he2022masked, tong2022videomae, huang2022masked} have shown remarkable performances in learning high-quality visual/audio representations with a high masking ratio from single-modality. These models mask the original input with a high ratio (\textit{e.g.} $75\%$) following different masking strategies (\textit{e.g.} random, grid), and only use the remaining input for feature extraction. High-quality feature representations are learned through the reconstruction strategy during model training. Afterward, the well-trained encoder will be used for further downstream tasks. 
However, these single-modality models learn representations only from the visual domain, whose ability will be limited when visual features are difficult to learn. For example, videos captured by security cameras in public areas, some real-world vlogs, and old videos archived from past decades are always blurry and of low quality, making it challenging for video understanding. 
Therefore, there remains a challenge: \textit{How to enhance video representation effectively and efficiently from multi-modalities?}

In reality, such single-modality methods have been extended into a multi-modality approach but require large computation resources. 
Gong \textit{et al.}
 have proposed CAV-MAE  \cite{gong2022contrastive} that learns joint representations from audio-visual modalities. However, to learn joint features, their proposed model uses three encoders (one encoder for each modality and one joint encoder) and reconstructs \textbf{both video and audio}. This strategy introduces added computational burden and complexity, diluting the focus on crucial features and introducing the potential for error propagation between modalities. 
 It also poses significant challenges due to the difficulty of synchronizing audio and video features,  degrading performance on certain downstream tasks such as action classification.
 In addition, CAV-MAE focuses on contrastive learning for feature alignment, 
 which highly relies on data quality and data augmentation \cite{he2022masked}.

Based on this, we propose AV-MaskEnhancer that learns high-quality video representations efficiently from both visual and audio modalities. 

\begin{figure*}[htbp]
  \centering
  \includegraphics[width=1.0\linewidth]{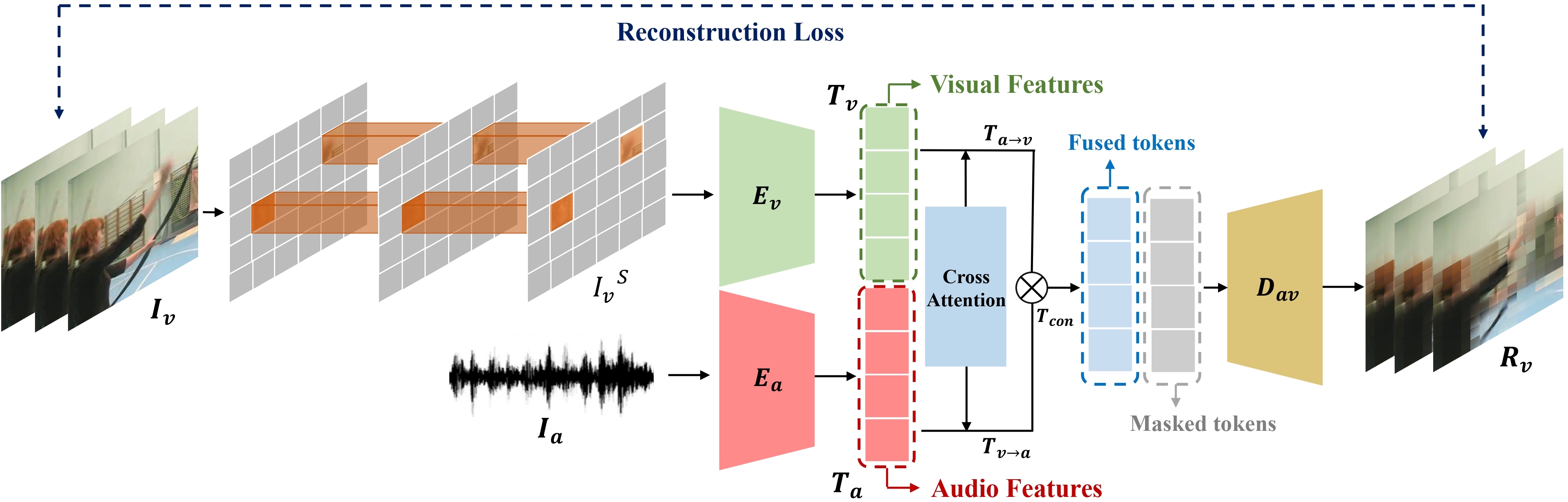}
  \caption{
    \textbf{The structure of our proposed AV-MaskEnhancer}. The orange color tube indicates \textit{tube masking} \cite{tong2022videomae} on the original input. The \textit{Visual Encoder} mainly follows VideoMAE \cite{tong2022videomae}, and the \textit{Audio Encoder} mainly follows the structure delineated by Tao \textit{et al.}\cite{tao2021someone}. Post encoding,  features are concatenated via cross-attention module \cite{tao2021someone} from visual and audio modalities, and the video is reconstructed based on this fusion. After training, the encoder is used for downstream video classification task. 
  }
  \label{fig:architecture}
\end{figure*}

In summary, our contribution is threefold:
\begin{enumerate}
    \item {
    We propose \textbf{AV-MaskEnhancer}, a novel framework efficiently and effectively leveraging audio-visual cross-modality for video reconstruction. For training the model, AV-MaskEnhancer uses two encoders to extract features from masked input in visual and audio modalities, and then utilizes a cross-attention module to align the cross-modality features. Afterward, a decoder is applied to reconstruct the video from fused features. After training, 
    we evaluate the video representation quality through the downstream video classification task. 
    }
    \item {
    We validate the superiority of our \textbf{cross-modality} strategy over single-modality. The utilization of audio modality allows the model to extract detailed features when the input video is low-resolution and blurry. We prove the effectiveness of this multi-modality approach by showing the experimental result compared with the unimodal method (\textit{e.g.} VideoMAE) on video reconstruction and video classification. 
    The result proves that AV-MaskEnhancer 
    enhances the performance of downstream tasks, proving its strong capacity to learn high-quality video representations.
    }
    \item {
    Our model achieves \textbf{state-of-the-art} video classification accuracy on the UCF101 dataset \cite{soomro2012ucf101}, with \textit{top-1} and \textit{top-5 accuracy} of $98.8\%$ and $99.9\%$, respectively. 
    UCF101 contains low-quality videos captured in uncontrolled environments such as featuring camera movements and diverse lighting conditions. This makes it an ideal data resource for developing a robust model to learn high-quality video representations. 
    }
\end{enumerate}

\section{Related Work}
\subsection{Transformers}
Transformer \cite{vaswani2017attention} is proposed by only using the attention mechanism and dispensing with recurrence and convolutions entirely.
The attention mechanism mainly considers the similarity between a \textit{query} and multiple \textit{key-value pairs}, which allows the model to focus on different parts of the input and capture long-range dependencies. 
This method speeds up the computation and achieves remarkable results in Natural Language Processing (NLP). Meanwhile,  Dosovitskiy \textit{et al.} extend the work in computer vision and propose the ViT \cite{dosovitskiy2020image}. Initially, ViT splits the image into different patches and linearly embeds each of them.
Considering the 2-D structure of the image, ViT adds positional embeddings with image patch embeddings, which allows the model to consider the ordering and position information from flattened image input. Because of its efficiency and capability of extracting long-range dependent features, 
ViT has become the mainstream backbone and has been applied in various tasks.

\subsection{Self-supervised models}
To address the high cost of data annotation and labeling, self-supervised approaches \cite{he2020momentum, chen2020improved}
have garnered significant interest in computer vision. Recently, there has been rapid growth in unsupervised training, particularly with the use of contrastive learning \cite{wu2018unsupervised, oord2018representation, he2020momentum, chen2020simple} . However, this method strongly depends on data augmentation \cite{chen2020simple, chen2021exploring, grill2020bootstrap}. Therefore,  the autoencoding-based strategy \cite{vincent2008extracting} is proposed, including the ImageMAE \cite{he2022masked}.

\subsection{MAE-based models with single modality}
The models based on MAE \cite{he2022masked, tong2022videomae, huang2022masked} have shown a strong ability to learn representations efficiently in a single modality (visual/audio). Generally, these models mask the original input with a surprisingly high ratio (\textit{e.g.} $75\%$, $90\%$) and are trained to reconstruct the input. 

Although these MAE-based models show great power, they learn representations only from a single modality. This strategy has unavoidable limitations when the features from that modality are difficult to extract.

\subsection{MAE-based models with multiple  modalities}
To address the shortcomings of the unimodal models mentioned above, the MAE-based models with multiple modalities have been proposed. As the most cutting-edge method,  CAV-MAE  \cite{gong2022contrastive} learns joint representations from audio-visual multi-modal data. Unlike our approach, the more complicated CAV-MAE utilizes three encoders (two exclusive encoders and one joint encoder) and reconstructs both video and audio, which involves high computation and requires more resources to converge the model. Moreover, this model focuses on the contrastive learning approach to align the distribution of multi-modality features, which highly relies on data quality and data augmentation \cite{he2022masked}. Therefore, the issue of learning high-quality video representations efficiently and effectively remains unsolved.
Section \ref{videoclasssection} proves that our model reaches a higher classification accuracy under the same conditions as CAV-MAE.

\begin{figure*}[htp]
  \centering
  \includegraphics[width=1\linewidth]{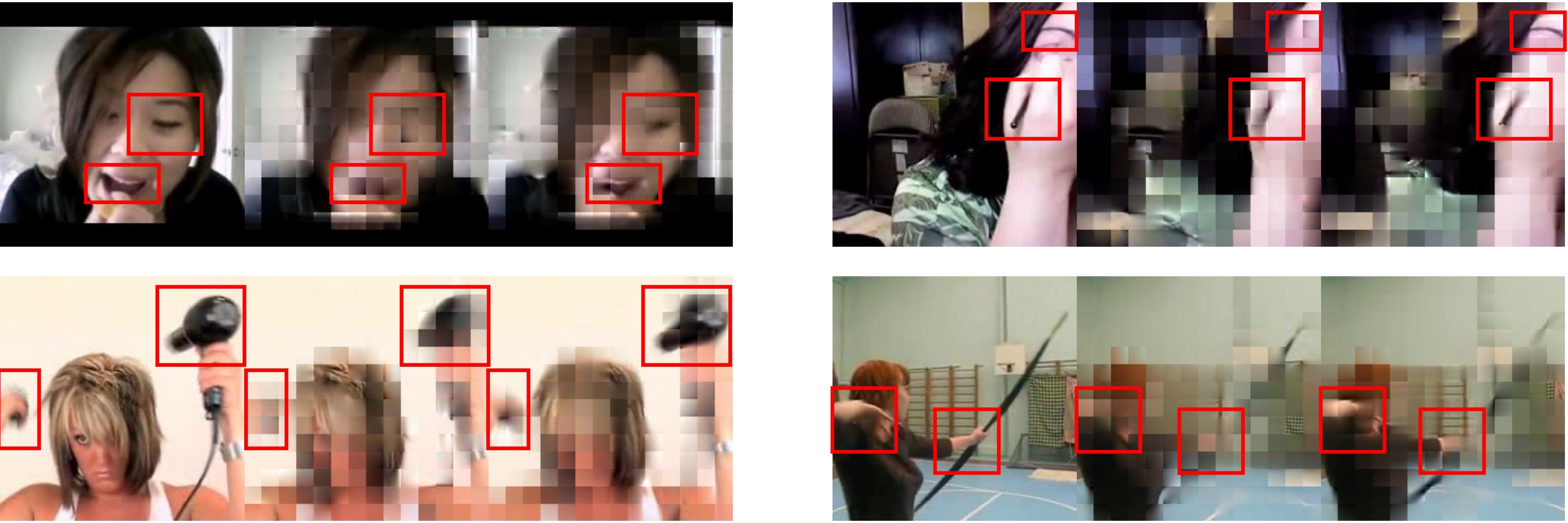}
  \caption{
    \textbf{The reconstruction results of UCF101 dataset}. In each pair of video clips, we demonstrate the \textbf{ground truth (Left)}, \textbf{VideoMAE \cite{tong2022videomae} (Middle)} result, and  \textbf{our (Right)} result. It is evident that our model can restore more details of the video clip compared with previous work, indicating the better representation the model has learned. 
  }
  \label{fig:duibitu}
\end{figure*}

\section{Method}
\subsection{Overall Architecture}
\label{overallarch}
Building upon the proven efficiency of the MAE-based structure in acquiring high-quality visual representations, our model adopts this encoder-decoder design as its foundational architecture to address the challenges mentioned previously. Our proposed AV-MaskEnhancer primarily extracts features from both video and audio inputs, and utilizes a cross-attention module for feature alignment. The model is trained through the video reconstruction strategy to learn a high-quality video representation. 

As shown in Figure \ref{fig:architecture}, our proposed AV-MaskEnhancer adopts two encoders ($E_v, E_a$) to extract deep features from visual and audio modality input ($I_v, I_a$), respectively, and one shared decoder $(D_{av})$ to reconstruct the video clip $(R_v)$ from fused tokens. 

Specifically, given the visual input $I_v$ with distribution as  $I_v \sim p_v(I)$ and audio input  as $I_a \sim p_a(I)$, deep feature tokens ($T_v, $ $T_a$) are extracted by encoders  $E_v$ and $ E_a$, respectively, as shown in Formula \ref{encoderformula}.
\begin{equation}
    T_v = E_v(I{_v}^{S}), T_a = E_a(I_a^{Cep})
\label{encoderformula}
\end{equation}
where $I{_v}^{S} = Sample(I_v)$ indicates the sampled input patches of $I_v$ because an extremely high masking ratio is applied. This considers the information redundancy in the visual modality mentioned in MAE-based models \cite{he2022masked}. Only the unmasked patches are fed into the visual encoder to learn visual modality representation. Meanwhile, $I_a^{Cep}$ is the vector of Mel-frequency cepstral coefficient (MFCC) in audio modality, which is derived from the original audio data using standard audio pre-processing techniques\cite{tao2021someone}.

To align the distribution of $T_v$ and $T_a$, a cross-attention module is adopted between these two modalities to obtain the new visual and audio tokens, $T_{a\rightarrow v}$ and $T_{v\rightarrow a}$, respectively. Afterward, the attention tokens are fused via concatenation and fed into the decoder for video reconstruction, as shown in Formula \ref{decoderformula}.
\begin{equation}
    R_v = D_{av}[Con(T_{v\rightarrow a}, T_{a\rightarrow v}), T_m]
\label{decoderformula}
\end{equation}
where $Con$ refers to the concatenation process and $T_m = E_v(I_v) - E_v(I_v^S)$ refers to the masked tokens of the original visual input $I_v$. 
This operation ensures that the size of the learned tokens matches that of the original video input.

\subsection{Visual Modality Structure}

Our proposed AV-MaskEnhancer structure in the visual modality primarily adheres to the VideoMAE structure \cite{tong2022videomae}, utilizing a ViT-based encoder. A notably high masking ratio of $90\%$ is employed on the input video to mitigate the temporal redundancy and visual information density. Consequently, only $10\%$ of $I_v$ is used to extract visual feature tokens.
The sampled video patches are then input into a ViT-based encoder to extract deep features, yielding tokens $T_v \in \mathbb{R}^{N_v\times D_v}$  from visual modality.

For each time, the model takes 
$N_{f}$ frames of video as input, and for each video frame, it is divided by $N_{p} \times N_{p}$ patches to be fed into the model. Additionally, each video is sampled for $N_{s}$ times. Therefore, we can get the first dimension of $T_v \in \mathbb{R}^{N_v\times D_v}$ as:
\begin{equation}
    N_v = N_{f} \times N_{p}^2 \times N_{s}
\end{equation}

Assuming each image patch has the size of $H  \times W$, and considering the RGB channels $C$, the second dimension of $T_v \in \mathbb{R}^{N_v\times D_v}$ can be expressed as:
\begin{equation}
    D_v = H \times W \times C
\end{equation}

Subsequently, $T_v$ is passed through a cross-attention layer. This layer generates attention features with $T_a$, which are used to fuse information and align the distribution between the visual and audio modalities.

\subsection{Audio Modality Structure}

In audio modality, $E_a$  is designed to learn audio representation by leveraging dynamic temporal information. Following \cite{tao2021someone}, the backbone of $E_a$ employs a 2D ResNet34-dilation network and is supplemented with a squeeze-and-excitation (SE) module \cite{chung2020defence}. Specifically, the dilated ResNet34 network aims to enable audio features $T_a$  to have the same dimension as $T_v$ for further concatenation.
As mentioned in Section \ref{overallarch}, the audio frame is firstly represented by $I_a^{Cep}$ prior to input into the model, which has the dimension of $I_a^{Cep} \in \mathbb{R}^{N_{len} \times N_{cep}}$, where $N_{len}$ indicates the audio length and $N_{cep}$ indicates the number of cepstrum extracted at each time unit. Afterward, $E_a$ takes $Cep$ as input and then generates $T_a \in \mathbb{R}^{N_a \times D_a}$ as the learned audio feature tokens, where $N_a$ is the number of audio frames and $D_a$ has the same size as $D_v$, as mentioned above.  The entire procedure in audio modality can be expressed as:
\begin{equation}
    T_a = E_a[Cep(I_a)]
\end{equation}

\subsection{Cross-attention Module}

Since the original visual and audio features are not well aligned along the time dimension, we apply a cross-attention module based on \cite{tao2021someone} which lets the audio feature contain video information, and vice versa. The cross-attention module aims to fuse visual-audio features and get well-aligned ones: $T_{a\rightarrow v}$ and $T_{v\rightarrow a}$. 
Taking visual features $T_{a\rightarrow v}$ as an example, we use audio features as attention query $(Q_a)$, and video features as key $(K_v)$ and value $(V_v)$.
Following the design proposed by \cite{tao2021someone}, the cross-attention module can be expressed as:
\begin{equation}
    \begin{split}
        T_{v\rightarrow a} &= V_a \cdot Softmax(\frac{Q_vK_a^T}{\sqrt{d}}) \\
        T_{a\rightarrow v} &= V_v \cdot Softmax(\frac{Q_aK_v^T}{\sqrt{d}})
    \end{split}
\end{equation}
where $Q, K, V$ indicate the vectors of query, key, and value, and $d$ denotes the dimension of these vectors \cite{tao2021someone}. After this operation, the new visual feature $T_{a\rightarrow v}$ contains the information from the audio modality, and the new audio feature $T_{v\rightarrow a}$ contains visual information correspondingly, implementing the alignment between these two modalities.

Through the cross-attention module, the new audio and visual features are temporally aligned and subsequently concatenated. Specifically, $T_{v\rightarrow a}$ and $T_{a\rightarrow v}$ are then concatenated along the time dimension to merge the information from these two modalities to obtain fused feature tokens  $T_{con} = Con(T_{v\rightarrow a}, T_{a\rightarrow v}) \in \mathbb{R}^{N_{fuse} \times D_{fuse}}$ where $N_{fuse}$ remains the same dimension as $N_v$ to further reconstruct the video and $D_{fuse} = 2 \times D_v$ because of the concatenation procedure. The concatenated feature tokens are then fed into decoder $D_{av}$ to generate videos.

\subsection{Decoder Structure}

The decoder design mainly follows the structure proposed in \cite{tong2022videomae}, where a lightweight ViT-based architecture is applied. Specifically, it has been proved that a decoder with 4 Transformer blocks yields the best result in video downstream tasks. Thus, we employ the same decoder architecture.

$D_{av}$ takes the concatenated feature tokens $T_{con}$ and masked tokens $T_m$ as input and reconstructs the video clip $R_v$, as shown in Formula \ref{decoderformula}.

\subsection{Masking Strategy}

Tong \textit{et al.} \cite{tong2022videomae} have proved the effectiveness of \textit{tube} masking, which expands a mask over the temporal axis, allowing different time frames to share the same masking map, as shown in the orange color \textit{tube} in Figure \ref{fig:architecture}. Therefore, we adopt this masking strategy in our paper, implementing an extremely high masking ratio of $90\%$ for each time frame.

\subsection{Loss Function}

Following the standard MAE-based methods \cite{tong2022videomae}\cite{he2022masked}, an MSE loss is employed to train the model. Specifically, an MSE loss is computed between the pixels of the original video $I_v$ and reconstructed video $R_v$ to compute the similarity. Given the original input video with distribution represented as $I_v \sim p_v(I)$, and the reconstructed video denoted as $R_v \sim p_v(R)$, MSE can be expressed as: 
\begin{equation}
    L_{MSE} = \mathbb{E}_{I_v \sim p_v(I), R_v \sim p_v(R)}[\frac{1}{n} \sum_{i=1}^{n} ((I_v)_i - (R_v)_i)^2]
\end{equation}
where $i$ refers to the $i^{th}$ pixel of $I_v$ and $R_v$, and $n$ refers to the number of total pixels.

\section{Experiments}

\subsection{Datasets}

We evaluate our AV-MaskEnhancer on the commonly used video dataset, UCF101 \cite{soomro2012ucf101}, which 
consists of over $13k$ ($9.5k/3.5k$ train/val) video clips across $101$ action classes and is grouped into five types: Human-Object Interaction, Body-Motion Only, Human-Human Interaction, Playing Musical Instruments, and Sports.

UCF101 contains web videos captured in uncontrolled settings, typically featuring camera movements, diverse lighting conditions, occasional partial occlusions, and occasional frames of low quality. This provides the condition to develop an advanced and robust encoder capable of learning high-quality video representations.

\begin{table*}[]
\centering
\fontsize{10}{12}\selectfont 
\begin{tabular}{c|cccccc}
\hline
\textbf{Methods} & \textbf{Backbone} & \textbf{Extra Data} & \textbf{Frames} & \textbf{Modalities} & \textbf{Top1-Acc}\\ \hline
XDC \cite{alwassel2020self} & R(2+1)D & Kinetics-400 & 32 & V+A & 84.2\% \\
GDT \cite{patrick2020multi} & R(2+1)D & Kinetics-400  & 32 & V+A & 89.3\% \\
MIL-NCE \cite{miech2020end}& S3D & HowTo100M$^*$ & 32 & V+T & 91.3\% \\
MMV \cite{alayrac2020self} & S3D-G & AS+HTM$^*$ & 32 & V+A+T & 92.5\% \\
CPD \cite{li2020learning} & ResNet50 & IG300k & 16 & V+T & 92.8\% \\
ELO \cite{piergiovanni2020evolving} & R(2+1)D & Youtube8M-2$^*$ & N/A & V+A & 93.8\% \\
XDC \cite{alwassel2020self} & R(2+1)D & IG65M$^*$ & 32 & V+A & 94.2\% \\
GDT \cite{patrick2020multi} & R(2+1)D & IG65M$^*$ & 32 & V+A & 95.2\% \\ \hline
VideoMAE \cite{tao2021someone}& ViT-B & Kinetics-400 & 16 & V & 96.1\% \\ 
CAV-MAE \cite{gong2022contrastive} & ViT-B & Kinetics-400 & 32 & V+A & 95.4\% \\ \hline
\textbf{AV-MaskEnhancer (95\% masked)} & ViT-B & Kinetics-400 & 16 & V+A & 96.4\% \\ 
\textbf{AV-MaskEnhancer (75\% masked)} & ViT-B & Kinetics-400 & 16 & V+A & 97.7\% \\ 
\textbf{AV-MaskEnhancer (90\% masked)} & ViT-B & Kinetics-400 & 16 & V+A & \textbf{98.8\%}\\ \hline
\end{tabular}
\begin{tablenotes}
       \small
       \item  1. * indicates the dataset is larger than Kinetics-400.
       \item 2. The Top-1 Accuracy, denoted as Top1-Acc, refers to the highest accuracy achieved in a video classification task. It is rounded to one decimal place for precision.
     \end{tablenotes}
\caption{\textbf{Comparison of AV-MaskEnhancer with State-of-the-art Methods on UCF101 Classification}. ``V" - visual modality, ``A" -  audio modality, ``T" - text modality, and ``N/A” - unavailable data. AV-MaskEnhancer outperforms all existing work and achieves state-of-the-art.}
\label{tab:comparison_with_multimodal_methods}
\end{table*}

\subsection{Video Reconstruction}

\begin{figure}[h]
  \centering
  \includegraphics[width=1.0\linewidth]{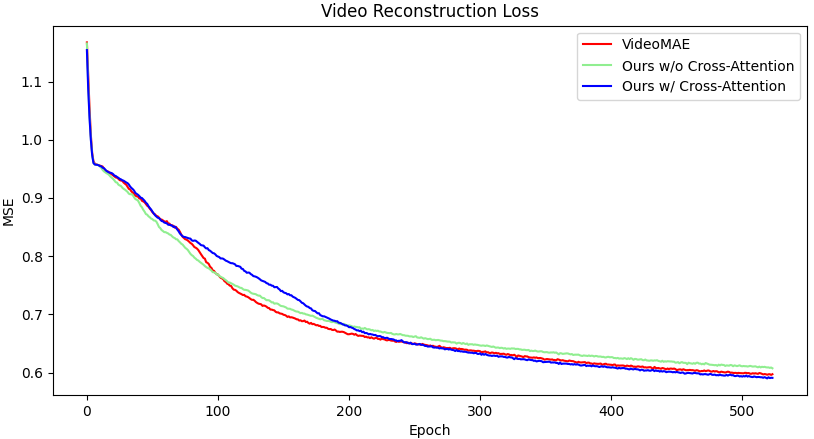}
  \caption{
    \textbf{Comparison of reconstruction loss during training}. We represent the loss of three models, charting them over the same epochs until converge. Our model with the cross-attention module has the lowest MSE loss, indicating the highest quality of the reconstructed video. 
  }
  \label{fig:line_plot}
\end{figure}

Following the standard MAE-based methods, we learn a high-quality video representation through reconstruction strategy. The video reconstruction result on UCF101 is shown in Figure \ref{fig:duibitu} and \ref{fig:line_plot}, proving the effectiveness of our cross-modality strategy and the cross-attention module. 

\subsubsection{Qualitative Result}
From Figure \ref{fig:duibitu}, it is evident to conclude that our model can restore more details of the original video, with the same masking ratio (90\%) as the previous work. The higher quality of the reconstructed video indicates that the model can learn a better representation of the visual feature, which is also proved in Section \ref{videoclasssection}.
Since the original video has a low resolution and does not provide much detailed visual information, it is difficult to reconstruct the missing pixels only from visual modality. 
Specifically, compared with previous work which only leverages visual information,  our model reconstructs better at the positions that make sounds. For example, our model restores the hairdryer better when the woman is drying her hair (shown in Figure \ref{fig:duibitu}). This result proves the effectiveness of our cross-modality approach. 

\subsubsection{Quantitative Result}
The effectiveness of the cross-attention module is quantitatively assessed by comparing the reconstruction loss (MSE) of VideoMAE and AV-MaskEnhancer with/without the cross-attention module, as shown in Figure \ref{fig:line_plot}. Despite our model does not converge the fastest during training ($\sim150$ epochs), it ultimately achieves the lowest reconstruction loss upon completion ($\sim500$ epochs), outperforming VideoMAE by reducing the MSE loss by $3\%$. This suggests the superior video clip quality produced by our model. The slower convergence of our model could be attributed to its usage of audio information for video reconstruction, which is a relatively more complex procedure.

\subsubsection{Cross-attention Module}
 We conducted an ablation experiment to test the classification accuracy on UCF101 with/without the cross-attention module. As shown in Table \ref{tab:avmae_with_without_att}, the utilization of a cross-attention layer improves the classification accuracy by $2.5\%$. 
The synchronization of audio-visual information provides valuable features for learning a joint representation, especially when the original audio and visual features aren't perfectly aligned. Thus, by leveraging the cross-attention module, our model targets the primary features that are common to both visual and audio modalities.

\noindent

\begin{table}[]
\centering
\begin{tabular}{cccc}
\hline
\textbf{Methods} & \textbf{Mask} & \textbf{Cross-attention} & \textbf{Top1-Acc} \\ \hline
AV-MaskEnhancer           & 90\%          & w/o                      & 96.3\%            \\
AV-MaskEnhancer           & 90\%          & w/                        & \textbf{98.8\%}   \\ \hline
\end{tabular}
\caption{\textbf{Top-1 accuracy of action classification on UCF101 by AV-MaskEnhancer}. The use of AV-MaskEnhancer in classification on UCF101 has resulted in an improved Top-1 accuracy (rounded to one decimal place), enhanced by $2.5\%$ by leveraging a cross-attention module.}
\label{tab:avmae_with_without_att}
\end{table}

\subsection{Video Classification}
\label{videoclasssection}
We conduct experiments on video classification of $101$ classes on UCF101  and compare our results with existing MAE-based methods including VideoMAE and CAV-MAE, as well as other existing state-of-the-art methods utilizing cross-modality strategy. Additionally, we present the results obtained using masking ratios of $75\%$, $90\%$, and $95\%$. We report the \textit{top-1 accuracy} as the evaluation metrics of the UCF101 classification task in Table \ref{tab:comparison_with_multimodal_methods}. 

We utilize weights pre-trained on the Kinetics-400 \cite{kay2017kinetics} dataset as extra data and incorporate them with our audio parts from the UCF101  to train AV-MaskEnhancer for around $1200$ epochs. Separately, we train our classification model using a ViT-B architecture and perform fine-tuning. 
Table \ref{tab:comparison_with_multimodal_methods} shows our results compared with other state-of-the-art multi-modality methods and MAE-based models, in which AV-MaskEnhancer outperforms all of the existing work and achieves the state-of-the-art, with \textit{top-1 accuracy}  as 98.8\%   and  \textit{top-5 accuracy} as 99.9\%.

In Table \ref{tab:comparison_with_multimodal_methods},  the comparisons between AV-MaskEnhancer and other multi-modality methods (\textit{e.g.} ELO \cite{piergiovanni2020evolving}, XDC \cite{alwassel2020self}, and GDT \cite{patrick2020multi}) show the effectiveness and the efficiency of our MAE-based structure, that our model can achieve higher accuracy even by using less extra data.  Moreover, although XDC  and GDT  use the same data (Kinetics-400) as ours, our model can achieve much higher accuracy, demonstrating the effectiveness of our model in enhancing representation learning.

Furthermore, the comparison between AV-MaskEnhancer and MAE-based models such as VideoMAE shows the success of involving audio modality. Since the original video has a low resolution, audio features can offer supplementary information to assist the model in better understanding the video content and consequently improving classification accuracy by $2.7\%$ with the same masking ratio ($90\%$). Even though VideoMAE has proved that $90\%$ masking ratio is the \textit{most effective} strategy \cite{tong2022videomae} for their model, we find that our approach achieves better classification accuracy than VideoMAE ($90\%$) even with a higher masking ratio of $95\%$. This outcome further reinforces the effectiveness of involving audio modality.

We also compare the results with the state-of-the-art model with audio-visual cross-modality, CAV-MAE \cite{gong2022contrastive}, under the same conditions. As shown in Table \ref{tab:comparison_with_multimodal_methods}, our model reaches a higher \textit{top-1 accuracy} which outperforms CAV-MAE by  $3.4\%$. Since CAV-MAE relies on contrastive learning to align the features from visual and audio modality, and it highly depends on data amount and data augmentation, it is reasonable that our model performs better by using the same/less data as CAV-MAE.

\section{Conclusion}

In conclusion, we propose AV-MaskEnhancer, a novel framework that leverages audio-visual cross-modality to learn high-quality video representation. Our model distinguishes itself by achieving state-of-the-art video classification accuracy on the UCF101 dataset, outperforming existing cross-modality work and the most cutting-edge MAE-based methods in both \textit{top-1} ($98.8\%$) and \textit{top-5} ($99.9\%$) accuracy. Through AV-MaskEnhancer, we demonstrate the superiority of our cross-modality strategy, enhancing video reconstruction and downstream task performance with the effective use of audio information.

\section*{Acknowledgment}

This work was extended from the Dartmouth College COSC 189 Video Understanding course, and was highly supported by the instructor Prof. SouYoung Jin from the Department of Computer Science, Dartmouth College, Hanover, US.

\bibliographystyle{plain} 
\bibliography{refs} 
\end{document}